\documentclass[twoside,11pt]{article}

\usepackage{jmlr2e}
\usepackage{xcolor}
\usepackage{makecell}
\usepackage{booktabs}
\usepackage{tcolorbox}
\PassOptionsToPackage{hyphens}{url}\usepackage{hyperref}
\usepackage{lastpage}
\usepackage{listings}
\usepackage{xspace}
\usepackage{pifont}

\makeatletter
\newcommand{\printfnsymbol}[1]{%
  \textsuperscript{\@fnsymbol{#1}}%
}

\definecolor{dkred}{rgb}{0.5,0,0}
\definecolor{dkgreen}{rgb}{0,0.6,0}
\definecolor{gray}{rgb}{0.5,0.5,0.5}
\definecolor{mauve}{rgb}{0.58,0,0.82}

\newcommand{\method}{{\sc{PyHealth} \xspace}}

\newcommand{\cmark}{\ding{51}}%
\newcommand{\xmark}{\ding{55}}%

\jmlrheading{21}{2021}{1-\pageref{LastPage}}{01/21; Revised
??/21}{??/21}{21-???}{Yue Zhao, Zhi Qiao, Cao Xiao, Lucas M. Glass, and Jimeng Sun}

\ShortHeadings{\method: A Python Library for Health Predictive Models}{Zhao, Qiao, Xiao, Glass, and Sun}

\firstpageno{1}

\begin{document}

\title{PyHealth: A Python Library for Health Predictive Models}

\author{\name Yue Zhao\thanks{Y. Zhao \& Z. Qiao contributed equally to this work.} \email zhaoy@cmu.edu \\
       \addr Carnegie Mellon University\thanks{Work initialized while at IQVIA.}\\
       \AND  
       \name Zhi Qiao\printfnsymbol{1} \email zhi.qiao@iqvia.com\\
       \addr Analytics Center of Excellence, IQVIA\\
       \AND
       \name Cao Xiao \email cao.xiao@iqvia.com\\
       \addr Analytics Center of Excellence, IQVIA\\
       \AND
       \name Lucas M. Glass
       \email lucas.glass@iqvia.com\\
       \addr Analytics Center of Excellence, IQVIA\\
       \AND
       \name Jimeng Sun
       \email jimeng@illinois.edu\\
       \addr University of Illinois Urbana-Champaign\\
       }

\editor{xxx xxx}

\maketitle

\begin{abstract}
Despite the explosion of interest in healthcare AI research, the reproducibility and benchmarking of those research works are often limited due to the lack of standard benchmark datasets and diverse evaluation metrics. To address this reproducibility challenge, we develop \method, an open-source Python toolbox for developing various predictive models on healthcare data. 

\method consists of data preprocessing module, predictive modeling module, and evaluation module. The target users of \method are both computer science researchers and healthcare data scientists. With \method, they can conduct complex machine learning pipelines on healthcare datasets with fewer than ten lines of code.
The data preprocessing module enables the transformation of complex healthcare datasets such as longitudinal electronic health records, medical images, continuous signals (e.g., electrocardiogram), and clinical notes into machine learning friendly formats. 
The predictive modeling module provides more than 30 machine learning models, including established ensemble trees and deep neural network-based approaches, via a unified but extendable API designed for both researchers and practitioners. 
The evaluation module provides various evaluation strategies (e.g., cross-validation and train-validation-test split) and predictive model metrics. 

With robustness and scalability in mind, best practices such as unit testing, continuous integration, code coverage, and interactive examples are introduced in the library's development. \method can be installed through the Python Package Index (PyPI) or \url{https://github.com/yzhao062/PyHealth}. 
\end{abstract}


\section{Introduction}

With the advances of machine learning technologies and the tremendous improvements in digital healthcare systems, digitized healthcare data have demonstrated initial success in enhancing health decision making and healthcare delivery
~\citep{doi:10.1093/jamia/ocy068}. However, health data are usually noisy, complex, and have heterogeneous forms, yielding a wide range of healthcare modeling tasks. For example, health risk prediction based on sequential patient data~\citep{10.1093/jamia/ocaa074}, medical image-based disease diagnosis~\citep{10.1093/jamia/ocaa280}, risk detection based on continuous physiological signals. e.g., electroencephalogram (EEG) or electrocardiogram (ECG) and multimodal clinical notes (e.g., text and images). The complexity and heterogeneity of health data and tasks lead to the long overdue of a dedicated ML system for benchmarking predictive health models despite their high value in healthcare research and clinical decision making.

Several efforts have been made to help approach these problems, including releasing benchmark urgent care datasets MIMIC-III \citep{harutyunyan2019multitask}, streamlining the analytic pipeline \citep{miller_taylor_2017_999013}. However, a dedicated toolkit for predictive health is absent. To fill the gap, we design and implement \method---a comprehensive Python library that provides a full-stack machine learning toolkit for predictive health tasks, including data preprocessing training and prediction, and result evaluation.

\method has five distinct advantages. First, it
encapsulates more than 30 state-of-the-art predictive health algorithms, including both classical techniques such as XGBoost \citep{chen2016xgboost} and recent deep learning architectures such as autoencoders, convolution based, and adversarial based models (see Table \ref{table:algorithms}). Second, \method has a wide coverage and contains models for different data types, e.g., sequence data, image data, physiological signal data, and unstructured text data.
Third, \method includes a unified API, detailed documentation, and interactive examples across all algorithms for clarity and ease of use---executing complex deep learning models needs fewer than ten lines of code. Fourth, most models in \method are covered by unit testing with cross-platform, continuous integration, code coverage, and code maintainability checks. Last, parallelization is enabled in select modules (data preprocessing) for efficiency and scalability, along with fast GPU computation for deel learning models through \texttt{PyTorch}. 

\begin{table}[!hbt]
\centering
	\caption{More than 30 Healthcare AI Models are implemented in \method 0.0.6} 
	\resizebox{0.9\textwidth}{!}{
	\begin{tabular}{l|cccc|c} 
	    \toprule
		\textbf{Method} &
		\textbf{Sequence} &
		\textbf{Image} &
		\textbf{Signal} &
		\textbf{Text} &
		\textbf{Deep Learning}\\
		\midrule

        RandomForest \citep{breiman2001random}             & \cmark     &   \cmark  &  \cmark  &  \cmark  & \xmark  \\  
        XGBoost \citep{chen2016xgboost}                    & \cmark     &   \cmark  &  \cmark  &  \cmark  & \xmark  \\
        BasicCNN \citep{CNN}                              &  \xmark     &   \cmark  &  \cmark  &  \cmark  & \cmark        \\
        \midrule
        \midrule
        LSTM \citep{hochreiter1997long}                    & \cmark     &    \xmark       &    \xmark      &    \xmark      & \cmark        \\
        GRU \citep{cho2014learning}                        & \cmark     &       \xmark    &    \xmark      &    \xmark      & \cmark         \\  
        RETAIN \citep{choi2016retain}                      & \cmark     &    \xmark       &    \xmark      &    \xmark      & \cmark         \\ 
        Dipole \citep{ma2017dipole}                        & \cmark     &    \xmark       &    \xmark      &    \xmark      & \cmark         \\
        tLSTM \citep{baytas2017patient}                    & \cmark     &    \xmark       &     \xmark     &    \xmark      & \cmark         \\          
        RAIM \citep{xu2018raim}                            & \cmark     &     \xmark      &    \xmark      &    \xmark      & \cmark         \\  
        StageNet \citep{gao2020stagenet}                   & \cmark     &     \xmark      &    \xmark      &    \xmark      & \cmark         \\
        \midrule
        \midrule
        Vggnet \citep{Vggnet}                             &   \xmark         & \cmark    &  \xmark        &    \xmark      & \cmark         \\  
        Inception \citep{Inception}                       &   \xmark         & \cmark    &   \xmark       &    \xmark      &
        \cmark         \\  
        Resnet \citep{Resnet}                             &  \xmark          & \cmark    &  \xmark        &    \xmark      & \cmark         \\ 
        Resnext\citep{Resnext}                            &  \xmark         & \cmark    &   \xmark       &    \xmark      & \cmark         \\
        Densenet \citep{Densenet}                         &  \xmark         & \cmark    &   \xmark       &    \xmark      & \cmark      \\    
        Mobilenet \citep{Mobilenet}                       &  \xmark          & \cmark    &  \xmark        &    \xmark      & \cmark      \\          
        \midrule
        \midrule
        DBLSTM-WS \citep{DBLSTM-WS}                       &  \xmark          &  \xmark         & \cmark    &    \xmark      & \cmark         \\          
        DenseConv \citep{DenseConv}                       &  \xmark          &  \xmark         & \cmark    &    \xmark      & \cmark         \\  
        DeepRes1D \citep{DeepRes1D}                       &  \xmark          &  \xmark         & \cmark    &    \xmark      & \cmark        \\
        Autoencoder+BiLSTM \citep{SDAELSTM}               &   \xmark         &  \xmark         & \cmark    &    \xmark      & \cmark        \\
        KRCRnet \citep{KRCRNN}                            &  \xmark          &  \xmark         & \cmark    &    \xmark      & \cmark         \\ 
        MINA \citep{MINA}                                 &  \xmark          &  \xmark         & \cmark    &    \xmark      & \cmark         \\
        \midrule
        \midrule
        Bidirectional-GRU \citep{cho2014learning}                       &  \xmark          &  \xmark         &    \xmark      & \cmark    & \cmark         \\
        DR-CAML \citep{DR-CAML}                       &  \xmark          &  \xmark         &    \xmark      & \cmark    & \cmark        \\
        DCAN \citep{DCAN}                       &  \xmark          &  \xmark         &    \xmark      & \cmark    & \cmark         \\  
        JointLAAT \citep{JointLAAT}               &   \xmark         &  \xmark         &    \xmark      & \cmark    & \cmark        \\
        MultiResCNN \citep{MultiResCNN}                            &  \xmark          &  \xmark         &    \xmark      & \cmark    & \cmark         \\ 
		\bottomrule
	\end{tabular}}
	\label{table:algorithms} 
\end{table}

\section{Library Design and Implementation}
\begin{figure*}[!tp]
\caption{\method contains three modules to support various healthcare Predictive tasks
}
    \begin{center}    
	    \includegraphics[width=0.95\linewidth]{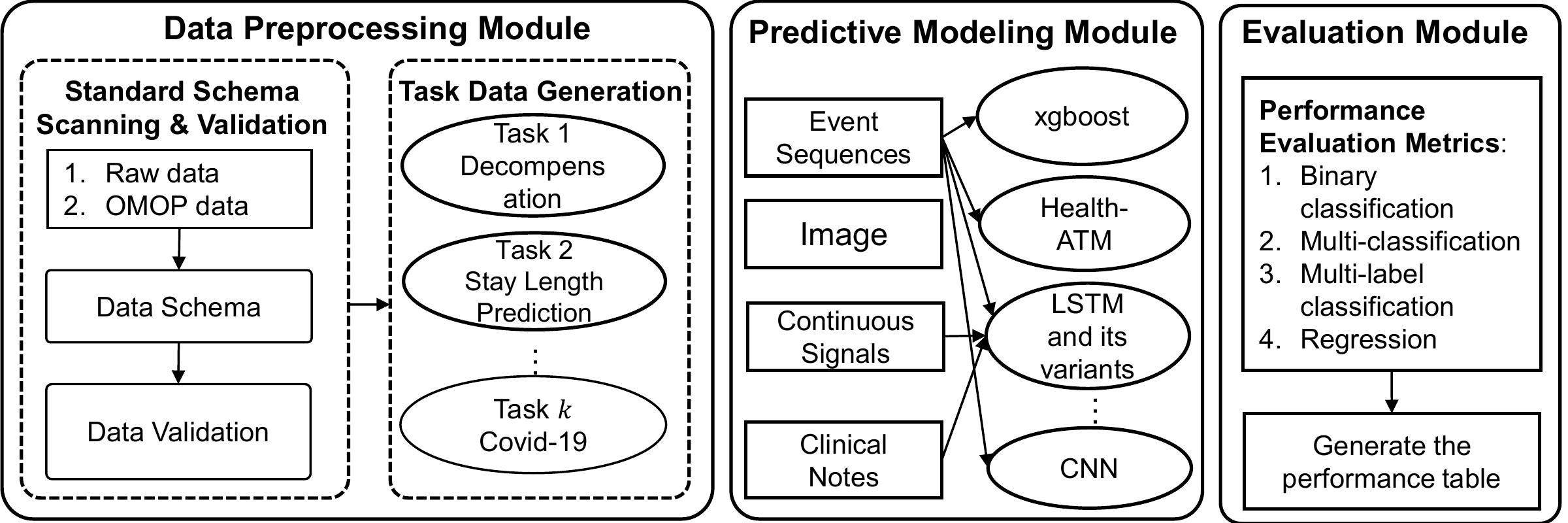}
	\end{center}
	\label{fig:flowchart}
\end{figure*}

\method is designed for Python 3 and relies on \texttt{numpy}, \texttt{scipy}, \texttt{scikit-learn} and \texttt{PyTorch}. As shown in Fig. \ref{fig:flowchart}, \method comes with three major modules: (i) \textbf{\textit{data preprocessing module}} can take user input and validate and convert the input data into the format that learning models can easily handle; (ii) \textbf{\textit{predictive modeling module}} comprises a suite of models that are grouped by input data type into \textit{sequence}s, \textit{image}, \textit{EEG}, and \textit{text}. For each data type, a group of dedicated learning models are implemented to be easily used, and (iii) \textbf{\textit{evaluation module}} can infer the task type automatically, e.g., multi-classification, and conduct the comprehensive evaluation by task type.

Inspired by \texttt{scikit-learn}'s API design and general deep learning design, most learning models inherit from the corresponding base class with the same interface: (i) \texttt{fit} processes the train data and validation data to learn the weights and save necessary statistics; (ii) \texttt{load\_model} selects the model with the highest validation accuracy and (iii) \texttt{inference} takes in incoming test data and make a prediction on it. A short demo of using these APIs is presented below. Other tasks follow a similar API.

\begin{lstlisting}[title={Code Snippet 1: Demo of PyHealth API with Mortality Prediction},captionpos=b]
  >>> from pyhealth.models.sequence.lstm import LSTM
  >>> from pyhealth.evaluation.evaluator import func

  >>> current_data = expdata_generator(exp_id=exp_id)   # initialize dataset
  >>> current_data.get_exp_data(sel_task="mortality")
  >>> current_data.load_exp_data()
  >>> model = LSTM(expmodel_id=expmodel_id, n_batchsize=20, use_gpu=True,     n_epoch=100)    # initialize LSTM model and set hyperparameters                          
  >>> model.fit(current_data.train, current_data.valid)                          
  >>> model.inference(current_data.test)      #  inference on test data
  >>> prediction_results = model.get_results()    # retrieve the test results
  >>> evaluation = func(prediction_results["hat_y"], prediction_results["y"]) 
\end{lstlisting}
\vskip 0.2in

Within the framework, a set of helper and utility functions (\texttt{check\_parameter}, \texttt{label\_check}, and \texttt{partition\_estimators}) are included in the library for quick data and model exploration. For instance, \texttt{label\_check} can automatically check the data label and infer the task type, e.g., binary classification and multi-classification.

\section{Project Focus}

\textbf{Build robustness}. We leverage continuous integration (e.g., \textit{Travis CI}, 
and \textit{CircleCI}) 
to conduct automated testing under various operating systems and Python versions. \\
\textbf{Community-based development and project relevance}. \method's code repository is hosted on GitHub\footnote{\url{https://github.com/yzhao062/PyHealth}} to facilitate collaboration. By the submission time, more than five people have contributed to the development and bug fix. \method has been used in various commercial projects at IQVIA, a leading healthcare research institute; researchers at the University of Illinois Urbana-Champaign and Carnegie Mellon University are actively using the library for algorithm design and benchmark.\\
\textbf{Documentation and examples}. Comprehensive documentation is developed using \texttt{sphinx} and \texttt{numpydoc} and rendered using \textit{Read the Docs}\footnote{\url{https://pyhealth.readthedocs.io/}}. It includes detailed API references, an installation guide, code examples, and algorithm benchmarks. \\

\section{Conclusion and Future Plans}
This paper presents \method, a comprehensive toolbox built-in Python for healthcare AI, with more than 30 classical and emerging learning models. A few future directions have been designed. First, we will enhance the library with specialized models for multimodal data. Second, the support of more different types of standard healthcare data formats will be provided. Additionally, we also plan to use \method to generate a large-scale benchmark for healthcare predictive tasks.


\clearpage
\newpage

\vskip 0.2in
\bibliography{ref}

\end{document}